\documentclass[aps,twocolumn,10pt,pra,superscriptaddress,amsmath,amssymb]{revtex4-1}

\usepackage{amssymb,amsfonts,amsmath}
\usepackage{graphicx}
\usepackage{color}
\usepackage{txfonts}
\usepackage[T1]{fontenc}
\usepackage[utf8]{inputenc}
\usepackage{CJK}
\usepackage{comment}
\usepackage{setspace}
\usepackage[usenames,dvipsnames]{xcolor}
\usepackage{soul}
\usepackage[normalem]{ulem}
\usepackage{listings}
\usepackage{hyperref}
\usepackage{type1cm}
\usepackage{lettrine}
\usepackage{courier}

\definecolor{dkgreen}{rgb}{0,0.6,0}
\definecolor{gray}{rgb}{0.5,0.5,0.5}
\definecolor{mauve}{rgb}{0.58,0,0.82}

\newcommand\abs[1]{\left|#1\right|}
\newcommand\etal{\textit{et al.}}
\newcommand\etc{\textit{etc.}}

\begin{document}
\begin{CJK}{UTF8}{mj} 

\title{Historic Emergence of Diversity in Painting: Heterogeneity in Chromatic Distance in Images and Characterization of Massive Painting Data Set}

\author{Byunghwee Lee}
\affiliation{Department of Physics, Korea Advanced Institute of Science and Technology, Daejeon, Korea}

\author{Daniel Kim}
\affiliation{Natural Science Research Institute, Korea Advanced Institute of Science and Technology, Daejeon, Korea}
\affiliation{Data Analytics Group, Samsung SDS, Seoul, Korea}

\author{Seunghye Sun}
\affiliation{Directorate of Culture and Arts Research, The Asia Institute, Seoul, Korea}
\affiliation{Cultural Cooperation Division, Ministry of Foreign Affairs, Seoul, Korea}

\author{Hawoong Jeong}
\email[Corresponding author; ]{hjeong@kaist.edu}
\affiliation{Department of Physics, Korea Advanced Institute of Science and Technology, Daejeon, Korea}
\affiliation{Asia Pacific Center for Theoretical Physics, Pohang, Gyeongbuk, Korea}

\author{Juyong Park}
\email[Corresponding author; ]{juyongp@kaist.ac.kr}
\affiliation{Graduate School of Culture Technology, Korea Advanced Institute of Science and Technology, Daejeon, Korea}
\affiliation{BK21 Plus Postgraduate Programme for Content Science, Daejeon, Korea}
\affiliation{Sainsbury Laboratory, University of Cambridge, Cambridge, United Kingdom}

\begin{abstract}
Painting is an art form that has long functioned as a major channel for the creative expression and communication of humans, its evolution taking place under an interplay with the science, technology, and social environments of the times. Therefore, understanding the process based on comprehensive data could shed light on how humans acted and manifested creatively under changing conditions. Yet, there exist few systematic frameworks that characterize the process for painting, which would require robust statistical methods for defining painting characteristics and identifying human’s creative developments, and data of high quality and sufficient quantity. Here we propose that the color contrast of a painting image signifying the heterogeneity in inter-pixel chromatic distance can be a useful representation of its style, integrating both the color and geometry. From the color contrasts of paintings from a large-scale, comprehensive archive of 179\,853 high-quality images spanning several centuries we characterize the temporal evolutionary patterns of paintings, and present a deep study of an extraordinary expansion in creative diversity and individuality that came to define the modern era.
\end{abstract}

\maketitle

\begin{figure*}[t!]
\includegraphics[width=0.9\textwidth]{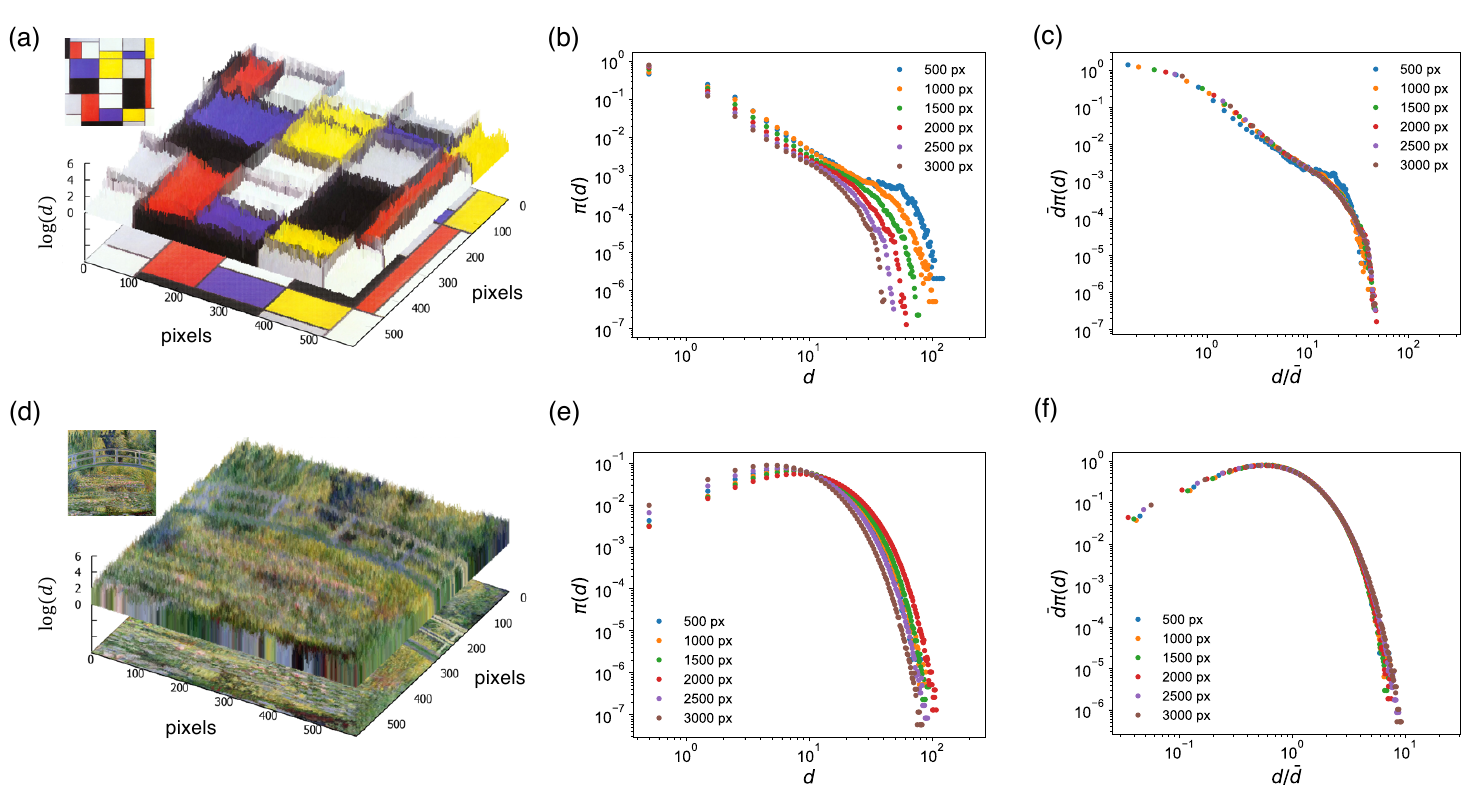}
\caption{Quantifying the color contrast of a painting from the color distances between adjacent pixels. The distance is visualized as height $d$ along the $z$-axis overlaid on the corresponding paintings, Piet Mondrian's \textit{Composition A} ((a)--(c)) and Claude Monet's \textit{Water Lilies and Japanese Bridge} ((d)--(f)). (a) In the Mondrian, a number of large $d$ correspond to the conspicuous walls between regular patches of uniform colors. (b) Such pattern can be shown in more detail via the distribution $\pi(d)$ (`o'), plotted in log-log scale. (c) The image size-dependent raw distributions can be rescaled into a single curve. (d) The Monet, meanwhile, lacks the crisp patchy structure of the Mondrian, indicative of heavily intertwining brushstrokes using complex color mixtures of the impressionism, resulting in high average $d$ but few extreme values. (e) The Monet's $\pi(d)$ accordingly shows a more rapidly decaying tail. (f) The distribution again collapses onto a single characteristic curve, regardless of image size. All images are obtained from Wiki Art and in the public domain.}
\label{interpixeldist}
\end{figure*}

\section*{Introduction}
Human have painted to express, record, and communicate ideas and recount experiences since long before the invention of writing~\cite{Gombrich1995}.  Painting thus has an essential and intimate connection to human history and, as a visual art form borne out of human sensitivity, imagination, and dexterity, is also a product of the human thought, science, and technology that determine the limits of what humans can envision and visualize on a physical medium such as a canvas. Such direct, intimate relationship between painting and science implies that a robust scientific study of painting could produce insights and reveal new answers to many pertinent questions in interdisciplinary field in quantitative and analytical manner.  To proceed with a scientific inquiry of paintings, we first establish that a piece of art can be viewed as a ``complex system'', as it is composed of heterogeneous elements that combine to effect novel emergent phenomena, a hallmark characteristic of one; in the case of an artwork, the stimulation of the senses the viewer experiences in its presence--be it cerebral, emotional, or physiological--cannot be attributed to a single element of it, for instance a single dot of a certain color, but the collective effect of all its parts. 

A recent development that is proving to have far-reaching implications for a scientific exploration of human actions and behavior in many social, cultural complex systems is the increasing availability of massive high-quality data that allows a large-scale application of scientific frameworks and verification~\cite{Manovich2015, WGA, BYP, WA, DPLA, GoogleArt}. In the area of culture, subjects on which quantitative pattern-finding have been performed to a degree include literature~\cite{Lutoslawski1897, Holmes2003, Binongo2003, Juola2006, Hughes2012} where Polish linguist Wincenty Lutos\l{}awski's work on the statistical features of word usage in Plato's Dialogue~\cite{Lutoslawski1897} is well known, music~\cite{Levitin2012, Manaris2005, Huron1991, Casey2008, Sapp2008}, and painting~\cite{Taylor1999, Taylor2007, Lyu2004, Hughes2010, DKim2014,Gatys2016, Graham2010, Elgammal2015, Rosa2015}. A landmark scientific study of paintings can be found in Taylor~\etal's characterisation of Jackson Pollock's (1912--1956) drip paintings using fractal geometry to distinguish between authentic Pollocks and those of unknown origins~\cite{Taylor1999}, demonstrating that an artistic style can be quantified.  More recent examples regarding painting include Lyu \etal's wavelet-based decomposition of images~\cite{Lyu2004}, Hughes~\etal's sparse-coding models for authenticating artworks~\cite{Hughes2010}, Kim~\etal's characterization of variations in \textit{chiaroscuro} technique via the so-called ``roughness exponent'' from statistical physics~\cite{DKim2014} and Gatys~\etal's style representation derived from correlations between the different features in different layers in a Convolutional Neural Network~\cite{Gatys2016}. Besides quantification of artistic styles, some studied perceived similarities between different paintings~\cite{Graham2010}, the influence relationships between artworks for quantifying creativity in an artwork~\cite{Elgammal2015}, and the changes in the perception of beauty using face-recognition on images from different eras~\cite{Rosa2015}.

Upon these progress in scientific analysis of painting, there still remains much necessity for a robust, comprehensive effort to overcome the following shortcomings therein: First, they often fall short of presenting a coherent and robust quantitative framework for analysis of multiple images; second, they do not use the full color information (due to the added complexity); third, they tend to focus on specific artworks or painters, not seeking generality, among others. In this work, we overcome these problems by formulating a framework for analyzing paintings that uses the complete color information which at the same time incorporates the geometrical relationships between the colors, two essential building blocks of an image. Our proposed quantity can be computed rapidly on the entire collection of digital images, allowing us to trace the stylistic evolution of painting throughout different periods, and identify significant patterns that characterizes each period.

Reflecting its ubiquity in nature and intriguing scientific characteristics, color boasts a long history as a subject of extensive scientific investigation in many fields such as physics (e.g., optics), biology (e.g., vision), and especially in the modern times, visual technology, to name only a few.  The beginning of modern quantitative research on color can be attributed to two groundbreaking investigations by Newton~\cite{Newton1979} and Goethe~\cite{Goethe1970,Ribe2002} who focused on the nature of light as the combinations of, and differentiations between, colors that lay foundations to more modern research on color and vision~\cite{Land1959,Land1977}. Inspired by these works and subsequent developments, here we propose the concept of `color contrast' as a signature of how color has been used in a painting. As its name suggests, color contrast refers to the compound effect of chromatic differentiation originating from different colors in a painting.  Well-known examples of paintings with intuitive, easily noticeable color contrast include Vincent van Gogh's (1853--1890) \textit{Starry Night} (1899) where a bright yellow moon is embedded in the dark blue sky and Piet Mondrian's (1872--1944) \textit{Composition A} (1923) where well-defined geometric shapes of distinct colors are juxtaposed to form the so-called `hard edge' painting, a style popularized during the twentieth century and became one of its signature styles. These two examples suggest that the sources of color contrast are the color difference (e.g., bright yellow versus dark blue) and the geometrical proximity (e.g., the juxtaposition of distinct colors generating clear, crisp boundaries).  Based on this realization, in this paper we devise a statistical measure of the color contrast in a painting we label \textbf{seamlessness} $S$, demonstrate that this quantity is indeed a useful indicator for characterizing distinct painting styles, and finally apply it to nearly $180\,000$ digital scans of historical paintings--the largest yet in our type of study--to track the evolution of painting and characterize how individual painters have developed creatively.

\begin{figure*}[t!]
\includegraphics[width=0.8\textwidth]{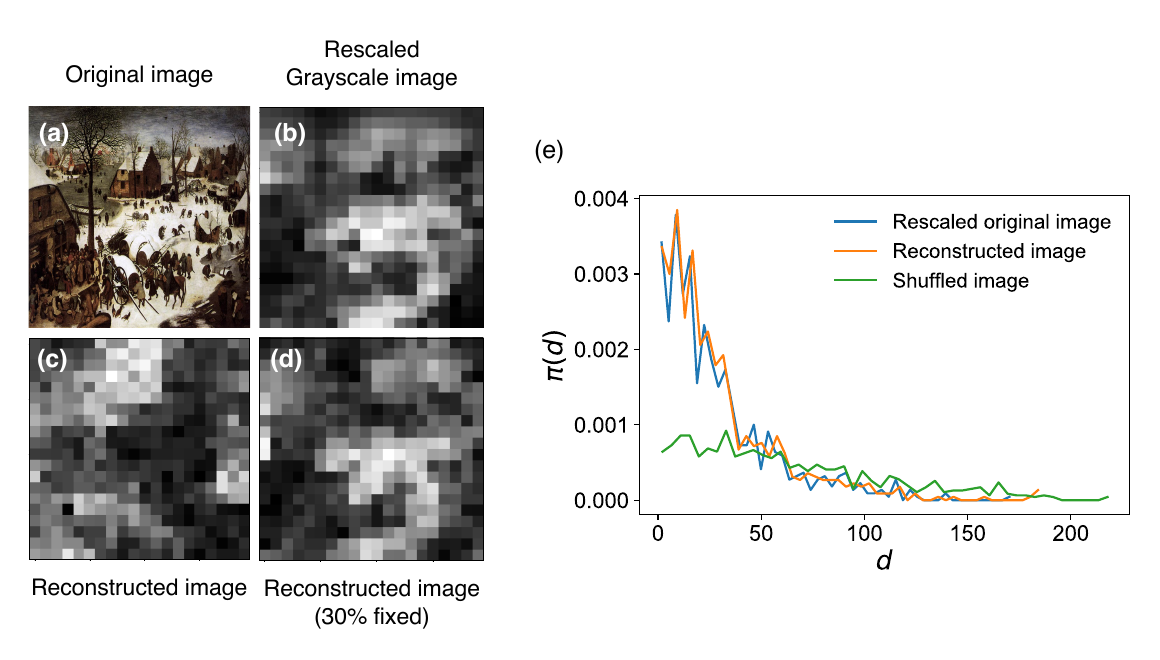}
\caption{Generating a reconstructed simulated image with the same inter-pixel color difference distribution as an input painting. (a) The input painting \textit{The Census at Bethlehem} by Pieter Bruegel the Elder (1566). The image is obtained from Wiki Art and in the public domain. (b) For a faster simulation we used a rescaled (20x20) grayscale version. (c) The reconstructed image from a completely randomized version of the original. While the locations of the patches of like colors have changed, they are of similar sizes as the original image. (d) The simulated image where 30\% of the pixels were maintained fixed in the original image. (e) $\pi(d)$ of the rescaled original image (b), reconstructed (c), and the randomly shuffled images.}
\label{ImgGeneration}
\end{figure*}

\section*{Data Description}
Digital scans of paintings (mostly western) were collected the following three major online art databases: Web Gallery of Art (abbreviated WGA)~\cite{WGA}, Wiki Art (WA)~\cite{WA}, and BBC-Your Paintings (BYP)~\cite{BYP}. The WGA contains paintings dated pre-1900, while the WA and BYP datasets contain those dated up to 2014 (all datasets are up-to-date as of Oct 2015).  WGA provides two useful metadata on the paintings: the painting technique (e.g., tempera, fresco, oil) and genre (e.g., portrait, still life, and `genre painting'--itself a specific genre depicting ordinary life). BYP is mainly a collection of oil paintings preserved in, and originating from, the United Kingdom. (We show that BYP data still exhibits a comparable trend in color contrast with other datasets.) The paintings dated pre-1300s were excluded, as they were too few.  Also excluded were those deemed improper for our analysis or outside the scope of it: They include partial images of a larger original, non-rectangular frames, seriously damaged images, photographs, etc. The final datasets used in our analysis contain 18\,321 (WGA), 70\,235 (WA), 91\,297 (BYP) images for a total of 179\,853. A significant majority of the images considered in this work--99.8\% of WGA, 76.0\% of WA, and all of BYP--are 500 pixels or larger in their length of longer side.

\begin{figure*}[t!]
\includegraphics[width=\textwidth]{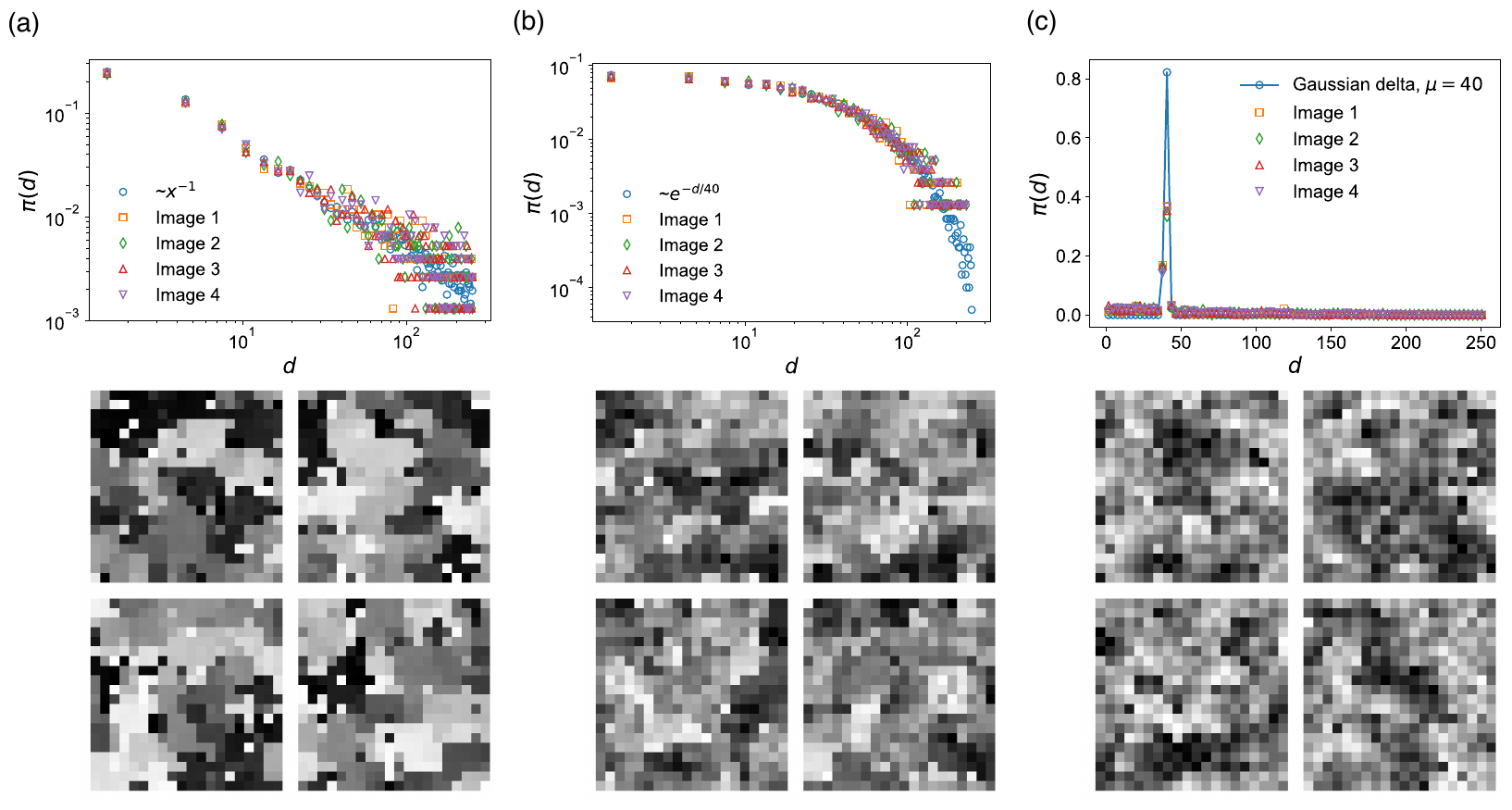}
\caption{Three distinct probability distributions $\pi(d)$ and simulated images. (a) Power-law distribution $\pi(d)\sim d^{-1}$ ($S=1$) with $\bar{d}=42$ (b) Exponential distribution ($\pi(d) \sim exp(-d/40)$) ($S=0$). (c) A narrow Gaussian distribution with mean $\bar{d}=40$ and $\sigma_d=1$ ($S\approx-1$). As we go from large $S$ (left) to small (right), the cluster of like colors become smaller, showing signs of lower color contrast.}
\label{RandomImg}
\end{figure*}

\section*{Results}
\subsection*{Characterizing color contrast of a painting from inter-pixel color difference distribution}
Color contrast represents the effect brought on by the differences in color between different points in a painting. It therefore can play a key role in characterizing the results of how a painter places different colors on a canvas in various positions, in other words, paintings. Human sense of color contrast between two colors in a painting (the pixels in case of a digital image) would be affected most strongly by two factors, the difference between the colors themselves and the geometrical separation---the more different the colors and the closer they are in real space, the more pronounced the effect of color contrast will be.  Quantifying color contrast with such a property thus requires two elements: A measure of the chromatic difference between two colors that agrees with human perception, and the spatial separation between the two.

Quantifying the difference between two colors starts by placing them on a three-coordinate system called `color space'.  A color space is named according to what the three coordinates measure. Commonly used ones include the RGB (Red, Green, Blue) space, the HSV space (Hue--position on the color wheel, Saturation, Value--brightness), and the CIELab space (the full nomenclature being 1976 CIE $L^*a^*b^*$) for $L^*$ (lightness between 0 for black and 100 for white), $a^*$ (running the gamut between cyan and magenta, but no specified numerical limits), and $b^*$ (between blue and yellow, similar).  To measure the color contrast we use the CIELab, as it was designed so that the human perception of the difference between two colors $(L_1^*,a_1^*,b_1^*)$ and $(L_2^*,a_2^*,b_2^*)$ would be proportional to the the Euclidean distance between the two, $d=\sqrt{(L_1^*-L_2^*)^2+(a_1^*-a_2^*)^2+(b_1^*-b_2^*)^2}$~\cite{Berns2000}. And in the present work, we take the simplest approach of considering the color distances between adjacent pixel pairs, which yields a total of $\sim 2N$ pixel pairs in a rectangular image of $N$ pixels to consider. Figs~\ref{interpixeldist} (a) and (d) visualize the differences between adjacent pixel colors for two paintings, Piet Mondrian's \textit{Composition A} and Claude Monet's \textit{Water Lilies and Japanese Bridge}.

We label the distribution of color difference between the $\sim 2N$ neighboring pixel pairs in a painting its `inter-pixel color difference distribution' $\pi(d)$. While the measured $\pi(d)$ is image-resolution dependent  (Figs~\ref{interpixeldist}~(b) and (e)), rescaling it by
\begin{equation}
\pi(d)=\frac{1}{\bar{d}}\mathcal{F}(\frac{d}{\bar{d}}), 
\label{scaling}
\end{equation}
where $\bar{d}=\sum_{d=0}^{\infty} d\pi(d)$ is the mean, caused distributions collapse into a single curve (Figs~\ref{interpixeldist}~(c) and (f)), demonstrating its size-independent universal characteristic. 

In Figs~\ref{interpixeldist}~(c) and (f), we see that the shapes of $\pi(d)$s from the two paintings are significantly different. In the Mondrian, a number of large $d$ correspond to the conspicuous walls between regular patches of uniform colors resulting in a heavy-tailed distribution of $\pi(d)$ compared to an exponential. The Monet, meanwhile, lacks the crisp patchy structure of the Mondrian, indicative of heavily intertwining brushstrokes using complex color mixtures of the impressionism, resulting in high average $d$ but few extreme values. The Monet's $\pi(d)$ accordingly shows a more rapidly decaying tail.

To see what types of painting a given $\pi(d)$ represents, we generate artificial images that possess the $\pi(d)$ of a real painting as input. The process starts by randomly relocating the pixels of the input image, then updating the image stepwise using the Metropolis-Hastings algorithm until the original painting's $\pi(d)$ is reconstructed, and inspecting the resulting image. To apply the Metropolis-Hastings algorithm we define the energy $E$ of an interim image $I$ to be the Kolmogorov–Smirnov (K-S) statistic between the $\pi(d)$'s of the interim image and the original
\begin{eqnarray}
	E(I)=\sup_x|\Pi_I(x)-\Pi(x)|,
\label{energy}
\end{eqnarray}
where $\Pi_I(x)$ and $\Pi(x)$ are the cumulative distributions of their $\pi(x)$, and $\sup_x$ denotes the supremum of the set of distances. The K-S statistic quantifies a distance between two cumulative distributions and is useful for nonparametric methods for comparing two sample distributions. Other statistical distances such as Jensen-Shannon divergence and Bhattacharyya distance may also be used for this purpose. Our Metropolis-Hastings process is as follows:  

\begin{enumerate}
\item	Initialize: The pixels of the original image are completely randomly shuffled, resulting in the initial configuration we label $I_0$.
\item	Generate a candidate configuration $I'$ by randomly choosing two pixels from the current configuration $I$ then switching their locations.
\item	Calculate the energy difference between $I$ and $I'$.
\item	Accept the new configuration with a probability

$ P(I\to I') =\begin{cases} exp(-\Delta E/T), & \text{if }\Delta E > 0 \\ 1, & \text{otherwise}.\end{cases}$.
\item	Proceed to next time step $t=t+1$, and repeat the processes $2-4$ until the target $\pi(d)$ is achieved.
\end{enumerate}

Temperature $T$ can be tuned to help escape local energy minima and help in convergence, and various techniques including simulated annealing could be employed to find approximate global energy minima~\cite{Kirkpatrick1983}. Fig~\ref{ImgGeneration} shows the method applied to Pieter Bruegel the Elder's \textit{Census at Bethlehem} (1566) and the final images obtained from using image generation process (see Figs~\ref{ImgGeneration}~(c) and (d)) using a reduced grayscale version (Fig~\ref{ImgGeneration}~(b)) for a faster simulation. The $\pi(d)$s of the original and the reconstructed image are shown in Fig~\ref{ImgGeneration}~(e). The reconstructed images using simulation, with identical $\pi(d)$, exhibits clusters of similar sizes and colors as the original, i.e. color contrast.  This does demonstrate that $\pi(d)$ indeed characterizes the color contrast of a painting. But $\pi(d)$ can be bothersome to use, so we devise a simpler measure derived from $\pi(d)$ itself, inspired by the relationship between the shapes of $\pi(d)$ and paintings shown in Fig~\ref{interpixeldist}.  The long- and short-tail distributions can be conveniently compared by the coefficient of variation $\sigma_d/\bar{d}$, where $\bar{d}$ and $\sigma_d$ are the mean and the standard deviation of $\pi(d)$.  A further desirable property of this quantity is that it is invariable under scaling of Eq.~\ref{scaling}. Other characterizing measures using higher moments of the distribution such as skewness or kurtosis also could be used as they are independent of location and scale parameters. The value of the coefficient of variation ranges between $0$ and $\infty$, $0$ for completely regular distributions such as a delta function ($\sigma_d=0$), 1 for an exponential or Poisson distribution ($\bar{d}=\sigma_d$), and $\infty$ for heavy-tailed distributions with an infinite variance.  For convenience, it is commonplace to use instead a quantity
\begin{equation}
S \equiv \frac{\sigma_d/\bar{d}-1}{\sigma_d/\bar{d}+1} = \frac{\sigma_d-\bar{d}}{\sigma_d+\bar{d}}
\end{equation}
which takes the range $[-1,1]$ of values. This quantity has found a wide range of use in various scientific fields, for instance in the study of inter-event time distributions such as analysis of earthquake occurrence patterns~\cite{Goh2008}, heartbeats of human subjects~\cite{Goh2008, Joh2012}, communication patterns of individuals~\cite{Wang2015}, and human behavioral dynamics online and offline ~\cite{Yasseri2012, Li2016}, etc. We do the same here, and we label this quantity the \textbf{seamlessness} of a painting, to be further explained below.  

In Fig~\ref{RandomImg} we show sample randomly generated grayscale images with $S$ taking the two extreme values and one the middle: (a) A power-law $\pi(d) \sim d^{-\alpha}$ with power exponent $\alpha=1$ ($S=1$), (b) an exponential $\pi(d) \sim exp(-\lambda d)$) with $\lambda=1/40$ ($S=0$), (c) a Gaussian distribution ($\bar{d}=40$) with a small width ($\sigma_d=1$) ($S\approx-1$).  In Fig~\ref{RandomImg}~(a), we see that the images with a power-law $\pi(d)$ (large $S$) exhibit interfaces of abrupt color change between extensive patches of similar colors to accommodate a large inhomogeneity in $d$, giving rise to a strong sense of overall color contrast.  Then in Fig~\ref{RandomImg}~(b) we see a weakened such effect: compared with (a), here the pixel-to-pixel color transitions are more gradual and relatively lack particularly sharp boundaries. Finally in Fig~\ref{RandomImg}~(c) we see a lack of sizable patches of uniform colors, resulting in blurred boundaries with a small $S$. This observations is also origin of our nomenclature `Seamlessness': A higher $S$ (Fig~\ref{RandomImg}~(a)) implies the image appears as if made up of a smaller number of patches (but each one being larger), requiring less seams (if one were to stitch them). A smaller $S$ (Fig~\ref{RandomImg}~(c)) means many smaller patches of different colors are intertwined, resulting in more seams. 

We further conduct a cluster size analysis on the simulated images to quantify our visual inspection. To do so we measure the color difference between adjacent pixel pairs (taking a value between 0 and 1 in a grayscale image) and link the pixels that are of 0.1 or a smaller value. Then the set of pixels that are connected via those links are considered to define a cluster of similar colors. We measure the size of the largest cluster and the average size of clusters to characterize each image. The generated images from the three different $\pi(d)$s in Fig.3 show quantitatively different characteristics. The largest cluster size of the images (whose full size is 20$\times$20), generated from a power-law distribution is 85.5 and the average cluster size is 7.01 on average (Fig~\ref{RandomImg}~(a)). The images following an exponential distribution (Fig~\ref{RandomImg}~(b)) have the largest cluster size as 62.75 and the average cluster size is 4.68 on average. Lastly, the largest cluster size of the images generated from a gaussian distribution is 11.0 and the average cluster size is 1.45 on average (Fig~\ref{RandomImg}~(c)). The difference in the size of largest clusters and the average cluster size of three distinct $\pi(d)$s shows that different $\pi(d)$s indeed exhibit different characteristics.

\subsection*{Mapping the Evolution of color Contrast from Massive Painting Data Sets}

\begin{figure*}[t!]
\includegraphics[width=0.9\textwidth]{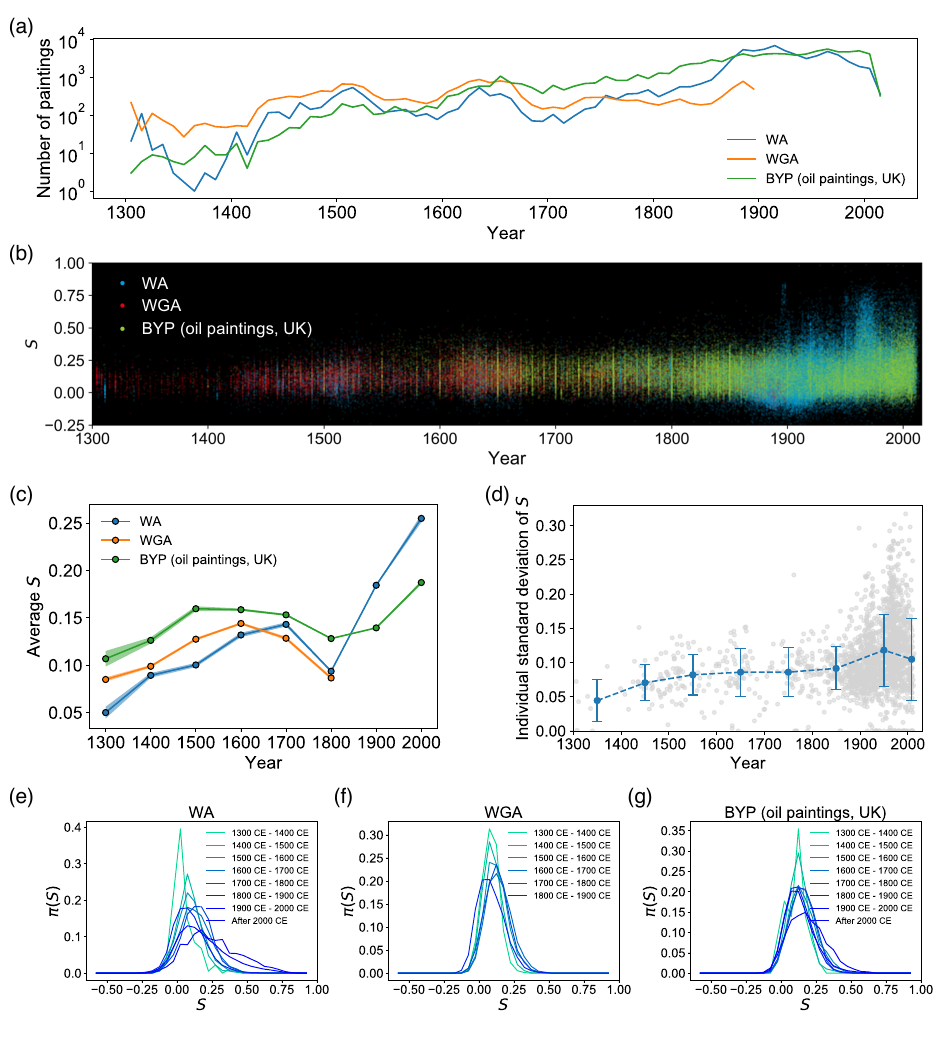}
\caption{The evolution of $S$ showing the development of paintings over time. (a) The number of paintings in the three datasets (WA, WGA, BYP) used in this study. (b) Scatter plot of S from 1300 CE to 2014 CE. We observe an  increase in the average and the variance of $S$, most noticeable in the mid-nineteenth century. (c) Changes in average S over time, along with the standard error of the mean. (d) Each individual painter’s standard deviation of S tends to grow, showing the widening diversity in style of works produced by a painter. Each gray dot indicates an artist. (e)—(g) The changing variances of S over time (WA, WGA, and BYP). The distributions become the broadest in the modern era. (The WGA dataset contains paintings only up to 1900.) }
\label{Evolution}
\end{figure*}

\begin{figure*}[t!]
\includegraphics[width=0.7\textwidth]{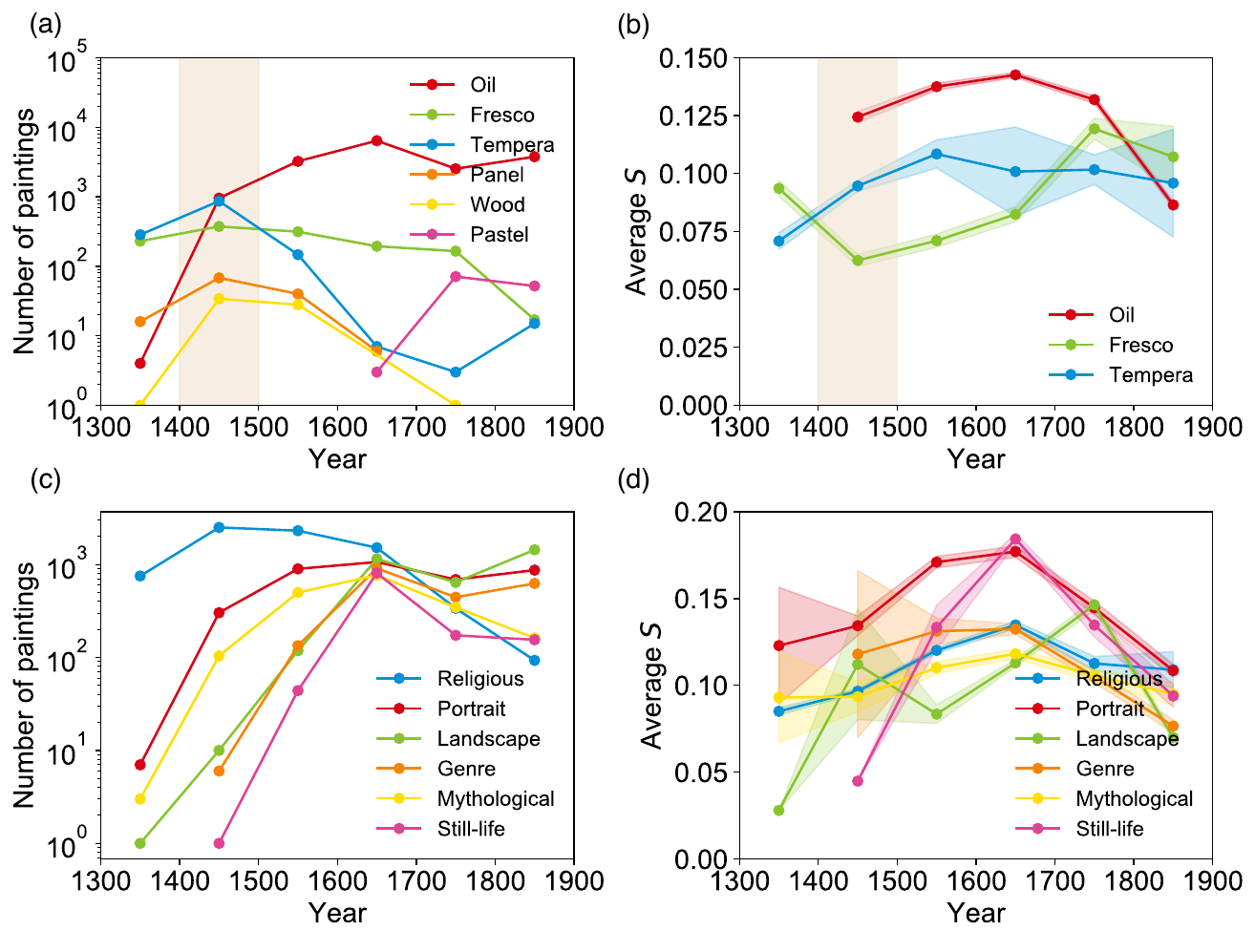}
\caption{Evolution of $S$ in paintings of various techniques and genres. (a) The number of paintings of various techniques in the WGA dataset. (b) Historical changes of $S$ in different painting techniques, with the standard error of the mean indicated. Kolmogorov-Smirnov tests on the shaded area confirm that the distribution of $S$ of different techniques are significantly different ($P<10^{-11}$ between every pair). (c) Number of paintings in various genres in the WGA dataset. (d) Evolution of $S$ in various genres, with the standard error of the mean indicated.}
\label{Technique}
\end{figure*}

$S$ measured from the data set is presented as a scatter plot in Fig~\ref{Evolution}~(b) with the date of production in the $x$-axis. Clearer statistical patterns of changes in color contrast are presented in Figs~\ref{Evolution}~(c)~to~(g). First, in Figs~\ref{Evolution}~(c)~to~(g), the average and standard deviation in $S$ have generally increased over time (with the exception of a temporary dip in the 18-19th centuries for the average $S$).  These changes can be found to correspond to notable and well-understood developments in painting technique and apparatuses.  For example, the increase in $S$ around the fifteenth century coincides with the adoption of oil as pigment binder medium (Fig~\ref{Technique}~(a))~\cite{Gombrich1995, Kleiner2016}; Before then, tempera (using egg yolk as binder medium) and fresco (watercolor painted directly on wet plaster; Michelangelo's Sistine Chapel ceiling painting is a famous example) were the most common.  The very physical characteristic of oil--high viscosity and the longer time to dry--provided painters with an opportunity to try new techniques that resulted in high contrast (Fig~\ref{Technique}~(b)), most notably \textit{chiaroscuro} (noted by gradations between dark and light that create the effect of highlighting the subject~\cite{Kleiner2016}) during the Renaissance period, and \textit{tenebrism} (representing a dramatic contrast between light and dark~\cite{Kleiner2016}) made popular during the Baroque period by such painters as Caravaggio (1517--1610).                                  

The emergence of such novel painting techniques is also closely related to the rise of novel painting genres: The ability to highlight the subject is credited for the rise in demand for portraits, for instance (Fig~\ref{Technique}~(c)). Still life, on the other hand, shows notable changes during the sixteenth century, reaching its peak in the seventeenth century (Fig~\ref{Technique}~(d)). The increase of $S$ in still life in the sixteenth century coincides with the changes in themes and subjects: In the first half of the century, Dutch painters such as Pieter Aertsen (1508--1575) and Joachim Beuckelaer (1533--1573) intentionally combined still life with detailed and bright depictions of biblical scenes in the background, while in the second half artists began to highlight still objects by incorporating chiaroscuro previously heavily used in portraits, resulting in high $S$~\cite{Kleiner2016}.

The most significant development occurred in the nineteenth century (Figs~\ref{Evolution}~(c)~and~(d)) when artists began to perceive paintings as a means of expressing one's individuality and originality more strongly than before~\cite{Eisenman1994}.  The pursuit of a wide range of different interpretations of the world gave rise to new techniques for expressing nature~\cite{Gombrich1995}.  In the beginning of the nineteenth century, the pursuit of fleeting impressions of light onto nature and landscapes replaced the dramatic, artificial lighting effect of the previous era, likely causing $S$ to drop.  The arrival of those ``impressionists'' were also helped by the railroad and the portable paints that enabled traveling to distant areas, leading to the surge in popularity of landscape paintings in the nineteenth century~\cite{Bomford1990} (Fig~\ref{Technique}~(c)).  Towards the end of the nineteenth century modern abstract art began to emerge, noted for an even more drastic departure from realism~\cite{Gombrich1995}.  After the decline in $S$ during the impressionist era, such simple and geometric abstraction led to a rapid increase in $S$ (Fig~\ref{Evolution}~(c)).  We note that, in addition, the increase in mean $S$ was accompanied by a significant increase in variance of $S$, indicating heightened diversity in style of paintings produced.  The most notable growth in variance occurs between the nineteenth and the twentieth centuries (Fig~\ref{Evolution}~(b)). Fig~\ref{Evolution}~(e) to (g) shows this in more detail: in earlier periods, the distribution of $S$ is narrow around the mean, but it becomes increasingly broader as we approach the modern times, rendering it less and less valid to talk of a `typical' style. Next we delve into the origin of this increased diversity in more detail.

\subsection*{Characterizing the individuality of painters in the modern era}
The patterns of $S$ shown in Fig~\ref{Evolution}~(e)~to~(g) are aggregate, i.e. over all the paintings contained in our data set. It thus cannot teach us about how varied the individual painters' styles may be, since two opposite explanations--painters having clear individual styles (therefore the heterogeneity coming from there being many different painters), or painters themselves exhibiting diverse styles--could lead to the same patterns. While in reality there would be both types of painters, we find that many modern painters have produced works that span a wide range of $S$, as shown in Fig~\ref{Evolution}~(d).  This culture of experimentation and embodiment of diverse stylistic possibilities are in good agreement with the characteristic of the modern era mentioned above~\cite{Gombrich1995}.  This prompts us to investigate the nature of individual stylistic diversity for the modern painter. Here we propose two distinct yet complementary aspects of stylistic individuality and explore them to better characterize the modern era, namely the individual painter's stylistic (1) evolution over their career that we call \text{metamorphosality}, and (2) uniqueness relative to the popular styles of the day that we call \textit{singularity}.

\begin{figure*}[t!]
\includegraphics[width=0.9\textwidth]{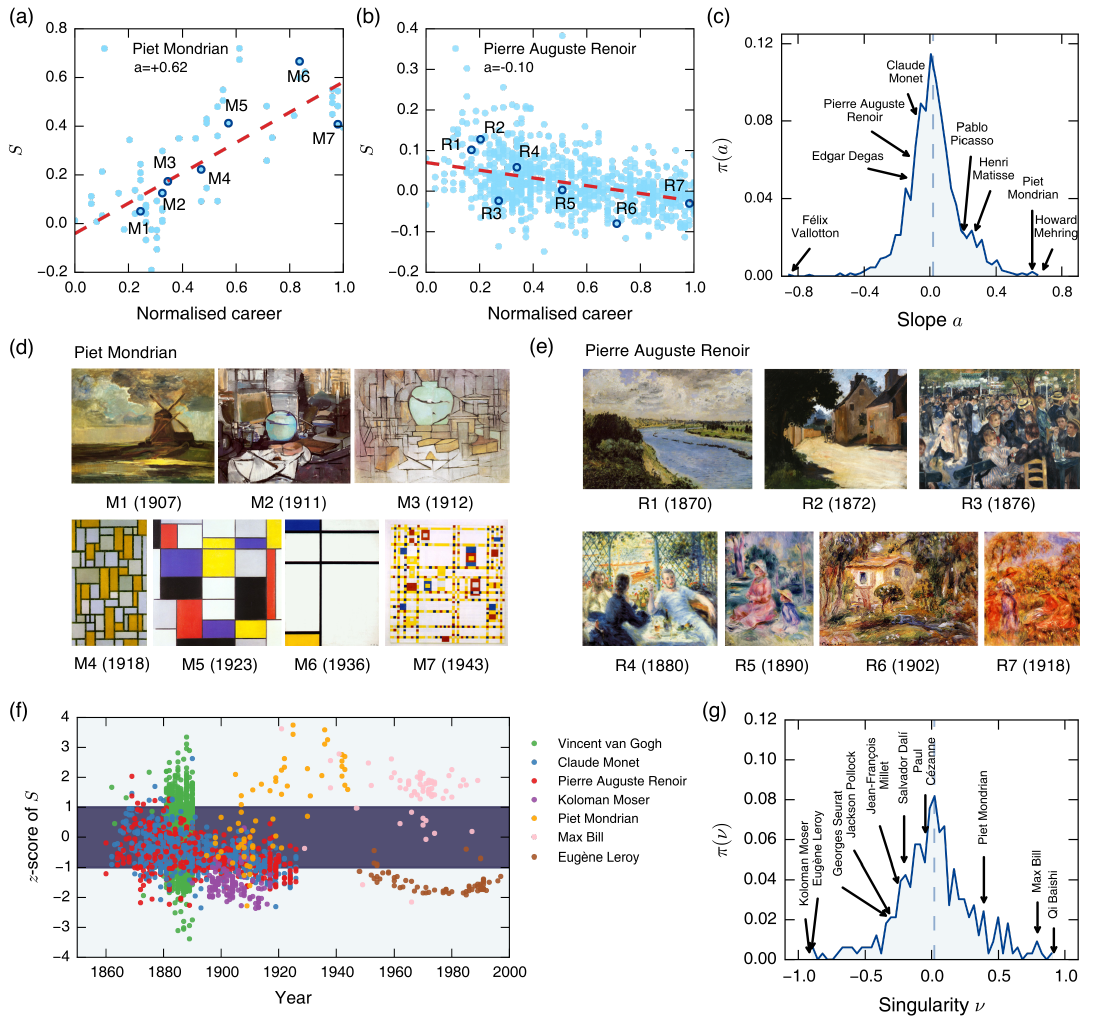}
\caption{Characterizing individual painters. (a, b) Growth in $S$ of Mondrian's and Renoir's paintings, respectively, over the normalized careers of each painter. The slope $a$ of the linear fit (dashed red lines) is $0.62$ for Mondrian and $-0.10$ for Renoir. (c) The histogram of the linear slopes $\{a\}$ of 1\,326 modern artists who produced paintings in at least five distinct years. A few notable artists are indicated. The dashed line indicates the average slope ($\bar{a}=0.02$, a slight trend towards abstract paintings). (d, e) Painting samples by Mondrian and Renoir, respectively, highlighting their stylistic changes over their careers (All images are obtained from Wiki Art and in the public domain). (f) Singularity of paintings by seven select artists. The darker band indicates the range $-1 \leq z \leq 1$. (g) Histogram of the singularity of 330 artists with more than 40 paintings. The dashed line indicates the average slope ($\bar{v}=0.02$). 
}
\label{Individual}
\end{figure*}

\begin{figure*}[t!]
\includegraphics[width=0.9\textwidth]{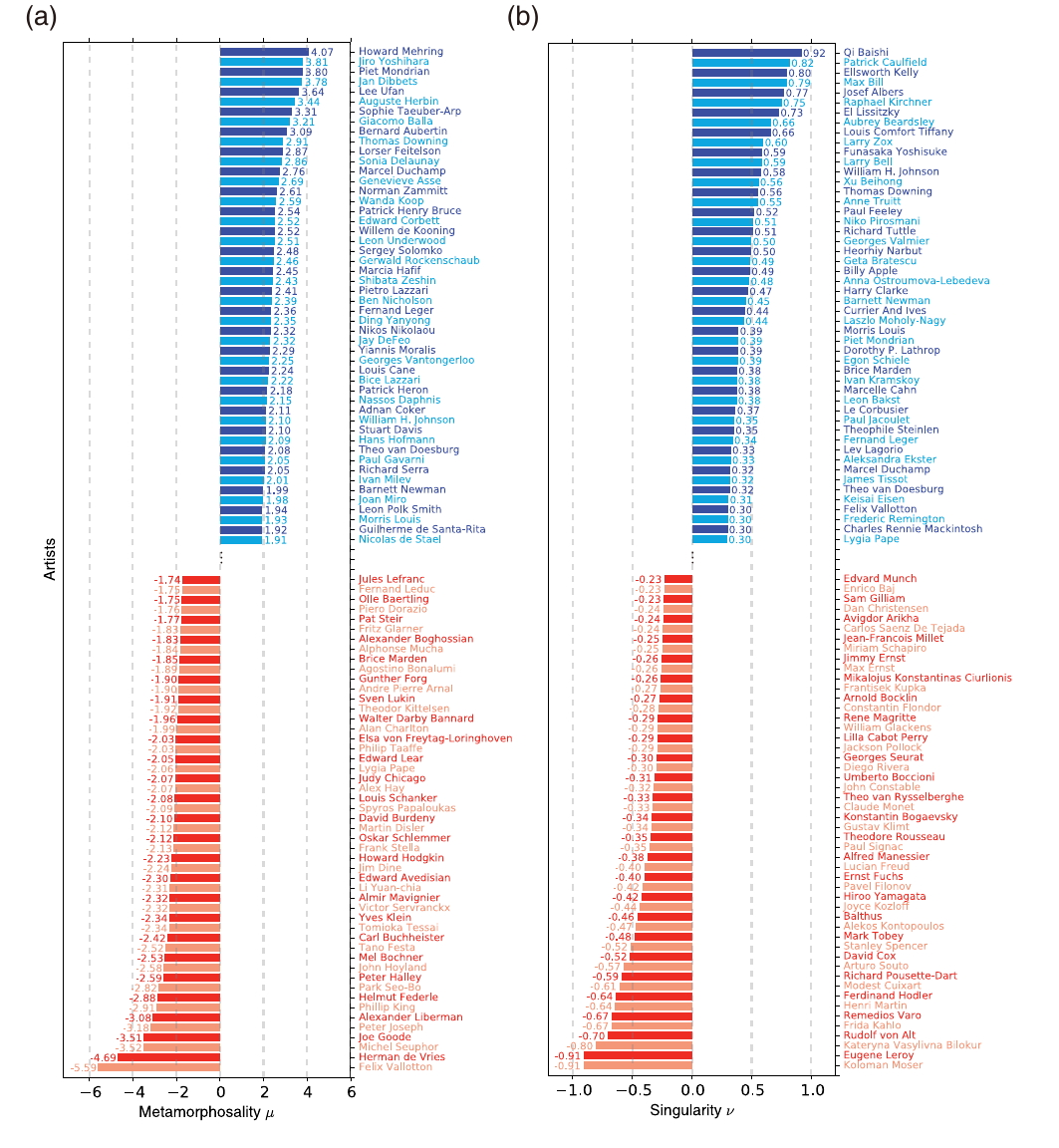}
\caption{Modern painters with the highest metamorphosality or singularity. (a) The 100 artists with the strongest metamorphosality $\mu$ (50 positively, 50 negatively). American painter Howard Mehring made the most significant shift from low-$S$ to high-$S$ during his career (top), while Felix Vallotton was the opposite (bottom). (b) The 100 artists with the strongest singularity $\nu$ (50 positively, 50 negatively). Qi Baishi's works contain the highest fraction of singularly high-$S$ paintings, while Kolomon Moser was the opposite.
}
\label{Bargraph}
\end{figure*}

\subsubsection*{Individual Evolution: Metamorphosality}
Mondrian, founder of \textit{De Stijl} movement and known for iconic abstractionism, in fact produced paintings that span a wide range of $S$ (Fig~\ref{Individual}~(d)). And it is reflected in how he progressed gradually from traditional style (small $S$) to abstractionism (large $S$) that matured in the 1920s (Figs~\ref{Individual}~(a) and (d)). Pierre Auguste Renoir (1841--1919), leader of early impressionism, exhibited the opposite trend: his $S$ decreases over time, as he transitions to more free-flowing brush strokes of impressionist techniques to generate boundaries that fuse softly with the background (Figs~\ref{Individual}~(b)~and~(e)). Other prominent impressionists such as Claude Monet (1840--1926) and Edgar Degas (1834--1917) demonstrate similar trends.

These observations prompt us to quantify such stylistic evolution of a painter using the rate of changes in $S$, given as the slope $a$ of the linear fit over one's career normalized to $1$. For instance, we find $a=0.62$ for Mondrian and $a=-0.10$ for Renoir (Fig~\ref{Individual}~(c)). The distribution of $a$ for the 1\,326 modern painters whose median of the production year is 1800 or later, (who produced paintings in five or more distinct years) resembles a Gaussian. Given this observation and that the quality of an artist is more reasonably measured in relation to others (as an absolute measure of artistic quality is not readily available), we define the \textit{metamorphosality} $\mu$ of a painter as the $z$-score $\mu\equiv(a-\bar{a})/\sigma_{a}$ of the painter's $a$, where $\bar{a}$ is the average, and $\sigma_{a}$ is the standard deviation. In Fig~\ref{Bargraph}~(a) we show the top 100 artists in terms of metamorphosality, fifty with increasing $S$ fifty with decreasing $S$.  On the positive side the American painter Howard Mehring (1931--1978) shows the largest $\mu=4.07$.  Accordingly, Mehring's early works are reminiscent of such figures as Pollock or Mark Rothko (1903--1970) and Helen Frankenthaler (1928--2011), employing scattered colors with vague boundaries~\cite{Mehring2016}. His later works, on the other hand, begin to feature geometric compositions of vivid colors with abrupt transitions, similar to Mondrian's hard-edge paintings.  At the other extreme with the most negative $\mu$ is Swiss-French painter F\'elix Edouard Vallotton (1865--1925), member of the post-impressionist avant-garde group Les Nabis. Initially having gained fame for wood cuts featuring extremely reductive flat patterns with strong outlines (high $S$), he produced classical-style paintings such as landscapes and still life in later life (low $S$) for $\mu=-5.59$.

\subsubsection*{Uniqueness Among Contemporaries: Singularity}
Another way to characterize a strong stylistic individuality would to measure how unique, or singular, a painting is.  It is again sensible to measure it in relation to other works, in this case especially among those made around the same time, since a style that is an outlier at one point in time may be mainstream at another, and vice versa. This can be achieved by computing the $z$-score of a painting's $S$ amongst its contemporary (defined as having been produced within five years of it).  We then call a painting highly singular if its $\abs{z}>z_c$, a threshold value which we set to be $1$ in this paper. In Fig~\ref{Individual}~(f) we show the $z$-scores of paintings of seven select painters as a scatter plot where those within the lightly-shaded areas represent the highly singular paintings ($\abs{z}>z_c$).  The figure teaches us that painters produced different ratios of highly singular works, indicating their conventional or unorthodox nature, and the styles they belong to (positive or negative $S$).  The \textit{singularity} $\nu$ of an artist is defined as the \textit{difference} between the fractions of their works in $z>1$ and $z<-1$.  Such definition of singularity give us the benefit of identifying those who tended to produce singular paintings and their preferred style (high or low $S$) simultaneously.  For example, 45\% of Mondrian's paintings are in $z>1$ (singularity high-$S$) and 6\% in $z<-1$ (singularly low-$S$) giving $\nu=0.39$, apparently consistent with his role in high-$S$ paintings. The histogram of the $\nu$ of 330 modern painters (who produced more than 40 paintings for sufficient data) of Fig~\ref{Individual}~(g) shows us the range of singularities among painters, including those even more singular than Mondrian. A more comprehensive list of the most singular painters (fifty for $\nu>0$ and fifty for negative) of Fig~\ref{Bargraph}~(b) contains many names who turn out to be highly regarded in fact for their  groundbreaking and unique styles: Examples include Qi Baishi (1864--1957), Chinese-born but very popular in the West for witty and vivid watercolors~\cite{COM2016}, has the largest singularity ($\nu=0.92$), followed by Max Bill (1908--1994) known for geometric paintings that came to symbolize the so-called `Swiss design' ($\nu=0.79$).  On the opposite side we find Koloman Moser (1868--1918), founding member of the Vienna Secession movement and known for complex repetitive motifs inspired by classical Greek and Roman art ($\nu=-0.91$), followed closely by Eug\`ene Leroy (1910--2000) known for numerous works featuring thick brush strokes in different colors, resulting in obscure and not readily identifiable imagery~\cite{Smith2016}, to name but a few.

\section*{Discussion}
This work presents a study to characterize the creative actions of humans from a massive, high-quality cultural data spanning several centuries up to the modern era. To accomplish it we devised a theoretical and computational framework for quantifying color contrast based on the relationship between the colors and geometry of the paintings.  We proposed quantifying the overall color contrast of a painting by the seamlessness statistic $S$ derived from the full distribution of the inter-pixel color differences, and using the Monte Carlo sampling methods from thermodynamics we demonstrated that $S$ is a consistent representation of color contrast.

Measurements of $S$ on the data were shown to capture in numerical terms multiple historically important developments (scientific, technological, technical, aesthetic, \etc) in that impacted the evolution of painting techniques, genres, and subjects. This has allowed us to present the stylistic evolution over human history and how it relates to the conditions of the times brought on by scientific, technological, cognitive innovations in a coherent and quantitative manner. 

To understand the greatly increased stylistic diversity of painting in the modern era, we profiled the individual qualities of painters using two criteria, metamorphosality (the variability of one's styles over a career) and singularity (uniqueness of style against one's contemporaries). We found that the stylistic diversity of painting in the modern era is due not to there being simply more painters, but to the emergence of painters actively evolving stylistically and producing original paintings that defied the established norms of the day.

We believe that our work shows a robust scientific methodology for modeling and analysis of complexity in visual artifacts using large-scale data. Our work could also be fruitfully applied to a variety of art forms which can clearly be converted to data representing its components and the relationship among them that allow us to find interesting patterns and information that can lead to new understanding of humans' creative process.

Based on our current investigation we can imagine multiple interesting directions for future research. First, more intricate analysis using $S$ would be possible and desirable in the immediate future to account for the possible biases across time and place due to the specific data set we used. Venturing further, non-western European or American art including Asian, Hindu, and Islamic painting art have been largely untouched in our work; large-scale analyses of these subjects would also be of immediate, universal interest. Also, integrating an analytical study using stylometric measures such as ours with object detection and segmentation techniques from machine learning could lead to a deeper understanding of art that incorporates both the styles and contents of paintings~\cite{Rosa2015, Krizheysky2012, He2017}. For example, how the same objects or motifs have been portrayed differently over time would shed light on changes in tastes as well as style. Going beyond the painting form, our work can also find use in understanding sculpture, architecture, visual design, film, animation, typography, \etc

\end{CJK}  

\end{document}